\documentclass[letterpaper]{article} 
\usepackage{aaai2027}  
\usepackage[hyphens]{url}  
\usepackage{graphicx} 
\usepackage{natbib}  
\usepackage{caption} 
\usepackage{algorithm}
\usepackage{algorithmic}
\usepackage{amsmath}
\usepackage{amssymb}
\usepackage{dsfont}
\usepackage{makecell}
\usepackage{tabularx}

\usepackage{newfloat}
\usepackage{listings}
\DeclareCaptionStyle{ruled}{labelfont=normalfont,labelsep=colon,strut=off} 
\floatstyle{ruled}
\newfloat{listing}{tb}{lst}{}
\floatname{listing}{Listing}

\usepackage{booktabs}
\usepackage{multirow}

\newcommand{\bboxresp}{\texttt{<bbox>}}
\newcommand{\absentresp}{\texttt{<absent>}}
\newcommand{\bboxformat}{\texttt{<bbox>}\allowbreak\texttt{[[x1,y1,x2,y2],...]}\allowbreak\texttt{</bbox>}}
\newcommand{\bboxmath}{\text{\ttfamily <bbox>}}
\newcommand{\absentmath}{\text{\ttfamily <absent>}}

\newcommand{\na}{--}

\newcolumntype{M}{>{\raggedright\arraybackslash}p{2.34cm}}
\newcolumntype{N}{>{\raggedleft\arraybackslash}p{\dimexpr(\textwidth-3.06cm-30\tabcolsep)/14\relax}}
\newcolumntype{Z}{>{\raggedleft\arraybackslash}p{0.72cm}}
\title{Knowing Whether, Not Just Where:\\
Anchor-Probing Distillation for Open-Set Remote Sensing Visual Grounding}
\author{
Zhihao Chang,
Lei Zhai,
Shuyuan Yang,
Zhixi Feng
}
\affiliations{
changzhihao@xidian.edu.cn, zhailei@stu.xidian.edu.cn,\\
syyang@xidian.edu.cn, zxfeng@xidian.edu.cn
}

\begin{document}

\maketitle

\begin{abstract}

Open-set visual grounding in remote sensing requires models to determine not only where a queried object is located, but also whether the image provides sufficient evidence for the query at all. Existing closed-set grounding models, however, tend to return plausible boxes even for absent objects, blurring localization confidence with genuine Visual Support. We introduce AnchorProbe-Distill (APD), a teacher-student framework that recasts open-set grounding as Visual Support learning. APD builds on the insight that supported queries should induce coherent spatial anchors, whereas unsupported or mismatched queries tend to produce unstable or misleading localization patterns. APD-T therefore uses the Original Query as a Reference Anchor to interpret query perturbations: Semantic-Preserving Queries retain the annotated target, Near-Category Ambiguous Queries are resolved according to whether they form a reliable alternative anchor or collapse onto the Reference Anchor, and Far-Category Counterfactual Queries provide rejection supervision. This reference-aware reasoning is distilled into APD-S, enabling unified localization and rejection from a single image--query pair. Experiments show that APD enables the resulting model to reject absent targets while also improving its localization of present targets.
\end{abstract}


\section{Introduction}

Remote sensing image understanding is moving from fixed-category recognition toward open-ended vision-language grounding. Given an aerial or satellite image and a natural-language query, a multimodal model is expected to localize the referred target with a bounding box. This capability is useful in disaster assessment, maritime monitoring, urban analysis, and other scenarios where analysts describe targets with flexible language instead of a predefined category list. Recent remote sensing vision-language models and datasets have made substantial progress on grounding, conversation, and multimodal instruction \cite{liu2023llava,kuckreja2023geochat,zhou2024geoground,zhang2024earthgpt,luo2024sky}. 

However, most grounding protocols still assume that the queried object exists in the image. This closed-set assumption is brittle in open-world use. An image--query pair has \emph{Visual Support} when the image contains sufficient spatial evidence for localization; otherwise, the correct response is rejection. A user may ask for an absent object, a visually unsupported category, or a near-category object that is not the annotated target. Yet current remote-sensing grounding models are predominantly trained to return a location for every query. They may therefore attach an unsupported query to a salient region, a visually similar object, or a contextually related location. We refer to this failure as \emph{hallucinated localization}. It is especially consequential in remote sensing, where scenes contain dense small objects, repeated textures, weak boundaries, and visually similar categories.

\begin{figure}[t]
\centering
\includegraphics[width=\linewidth]{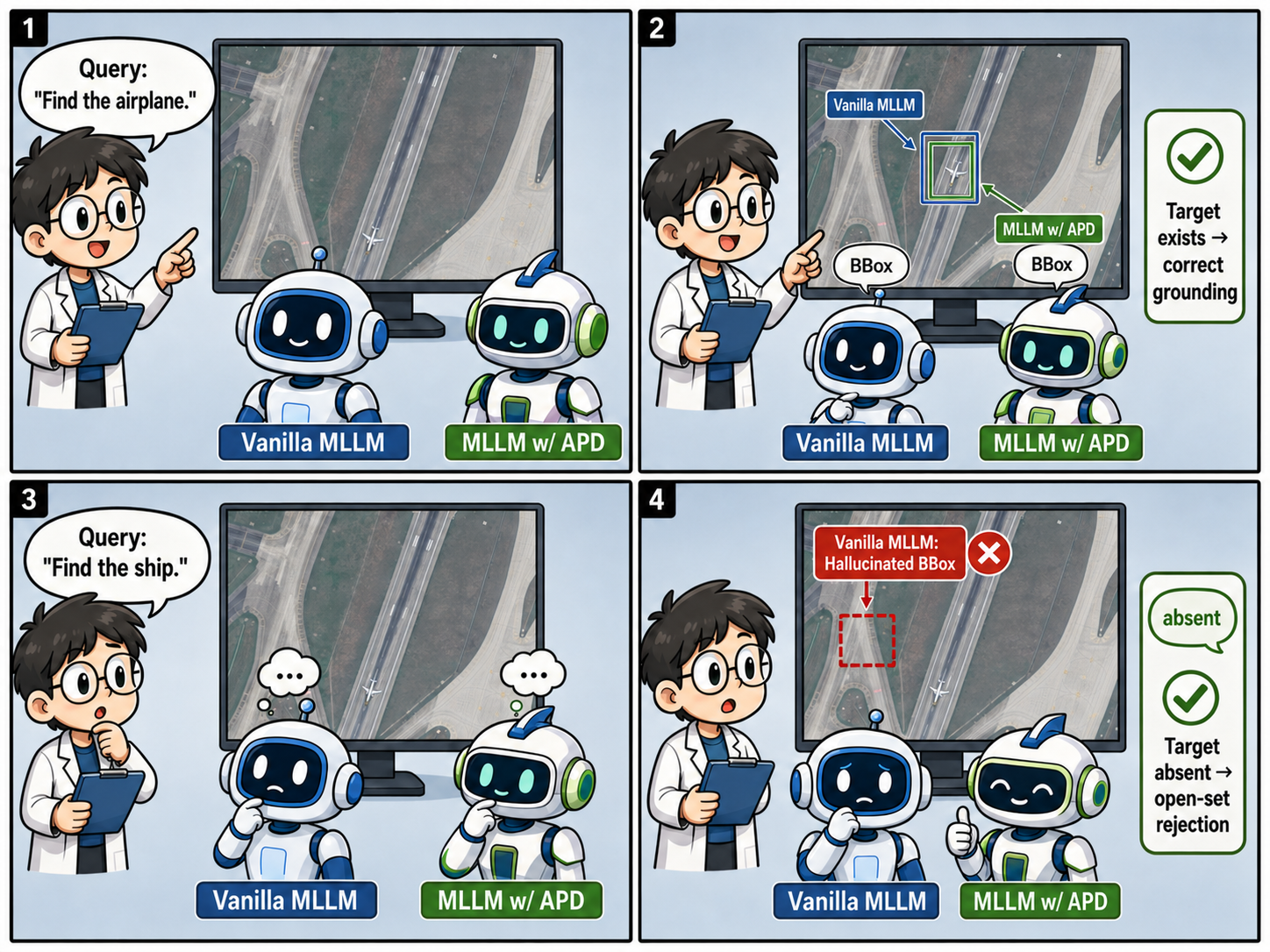}
\caption{Hallucinated localization. Both models localize the supported airplane query; for the absent ship query, a vanilla MLLM predicts a box while APD returns \absentresp{}.}
\label{fig:intro_motivation}
\end{figure}

\begin{figure}[t]
\centering
\includegraphics[width=\linewidth]{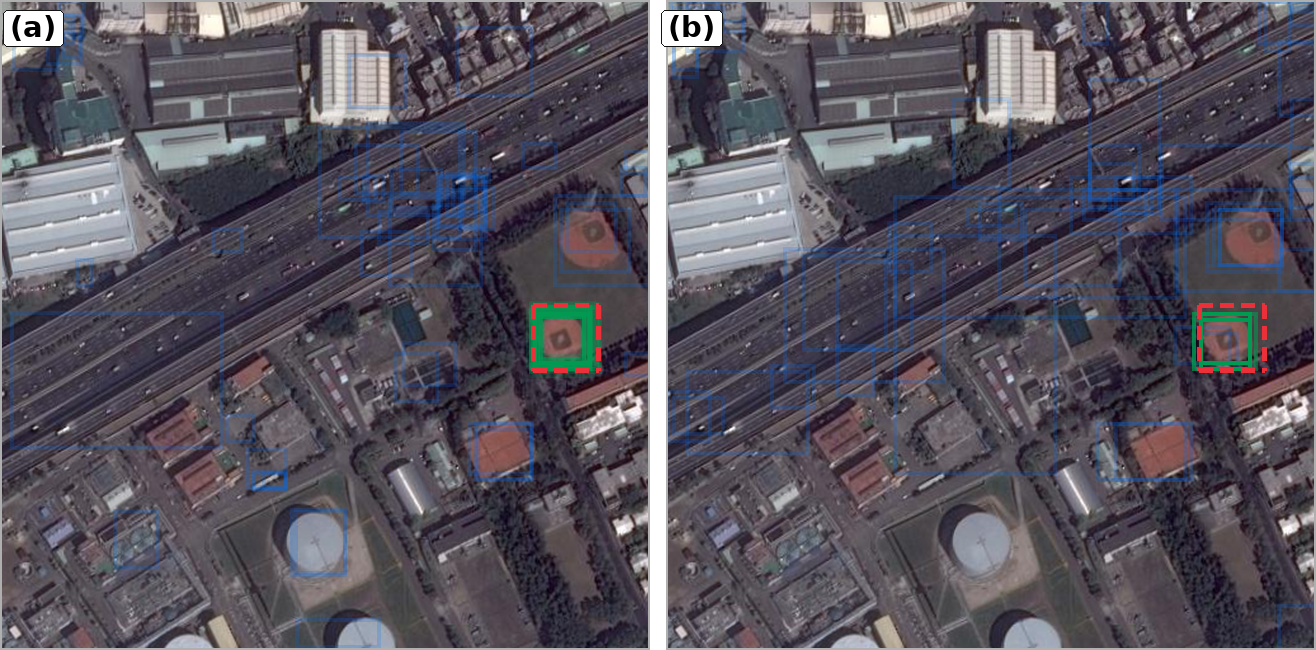}
\caption{Stochastic predictions from (a) Qwen3-VL-2B-FT and (b) GeoGround. Green and blue boxes denote the baseball-diamond and harbor queries, respectively; dashed red is ground truth.}
\label{fig:intro_sampling}
\end{figure}

As shown in Figure~\ref{fig:intro_motivation}, a closed-set grounding model can hallucinate a plausible location when the queried target is absent, a failure that recurs in Figure~\ref{fig:intro_sampling} under stochastic prediction. For both Qwen3-VL-2B-FT and GeoGround, the supported baseball-diamond query produces a compact group of boxes around the annotated target, whereas the unsupported harbor query still elicits valid-looking boxes at inconsistent or misleading locations. These examples indicate that a single plausible box does not establish Visual Support; a supported query should instead induce a coherent and trustworthy spatial anchor across repeated predictions. This observation directly motivates APD to probe localization stability and interpret each candidate anchor relative to a supported Reference Anchor. Directly prompting a model to output \absentresp{} remains unreliable because rejection competes with the category priors, contextual associations, and visual saliency learned during closed-set grounding. Manually collecting negative image--query pairs is costly, whereas automatically or randomly constructed negatives can be noisy.

We propose \textbf{AnchorProbe-Distill} (APD), an online teacher-student framework that converts Internal Stability and Consensus Ratio into type-aware supervision. Starting from existing positive grounding samples, APD-T repeatedly probes a teacher and exploits each Original Query as a reliable reference for interpreting perturbed queries. Semantic-Preserving Queries inherit the annotated box; Near-Category Ambiguous Queries become spatially distinct positives or overlap-based hard negatives only when their anchors pass reference-calibrated reliability tests; and Far-Category Counterfactual Queries provide explicit rejection targets. APD-S then distills these localization and rejection targets into a single-pass model, eliminating reference queries, repeated sampling, and consensus computation at inference time. Our main contributions are threefold:
\begin{itemize}
\item We formulate open-set remote-sensing grounding as Visual Support learning, distinguishing genuine grounding from hallucinated localization through the geometric stability of repeated predictions.
\item We introduce a teacher-student distillation paradigm in which APD-T converts reference-calibrated stochastic probes into type-aware localization or rejection targets, and APD-S preserves this verification capability during single-pass inference.
\item Experiments on RefGeo and its paired Far-Category set demonstrate consistent gains across multiple VLMs. Relative to fine-tuned Qwen3-VL-2B, APD-S improves Overall Acc@0.5 from 62.48\% to 66.30\% and RR from 0.05\% to 97.81\%.
\end{itemize}

\section{Methodology}

\subsection{Overview}

AnchorProbe-Distill treats open-set grounding as a problem of learning Visual Support. A closed-set grounding model can often produce a syntactically valid box even when the query is absent; therefore, the key supervision signal is not only where the object is, but whether the image provides enough spatial evidence to justify returning a box. APD implements this idea with a teacher-student design. APD-T serves as a teacher-side verifier that converts repeated predictions into Visual Support labels, and APD-S distills these labels for single-pass inference.

The method has three conceptual steps. First, APD-T converts stochastic teacher predictions into an anchor-level support estimate: a query is visually supported only when repeated predictions concentrate around a coherent geometry. For a perturbed or counterfactual query, the Original Query provides the Reference Anchor, so support is judged relative to the teacher's reliability on the same image rather than by an image-independent threshold. Second, APD-T interprets perturbations according to their semantic role. Semantic-Preserving Queries inherit the annotated target; reliable Near-Category Ambiguous Queries become spatially disjoint positives or overlap-based hard negatives according to their relation to the Reference Anchor; and Far-Category Counterfactual Queries provide negative supervision. Third, APD-S distills these online labels into a grounding model, removing the paired probing procedure at deployment.

\begin{figure*}[t]
\centering
\includegraphics[width=2\columnwidth]{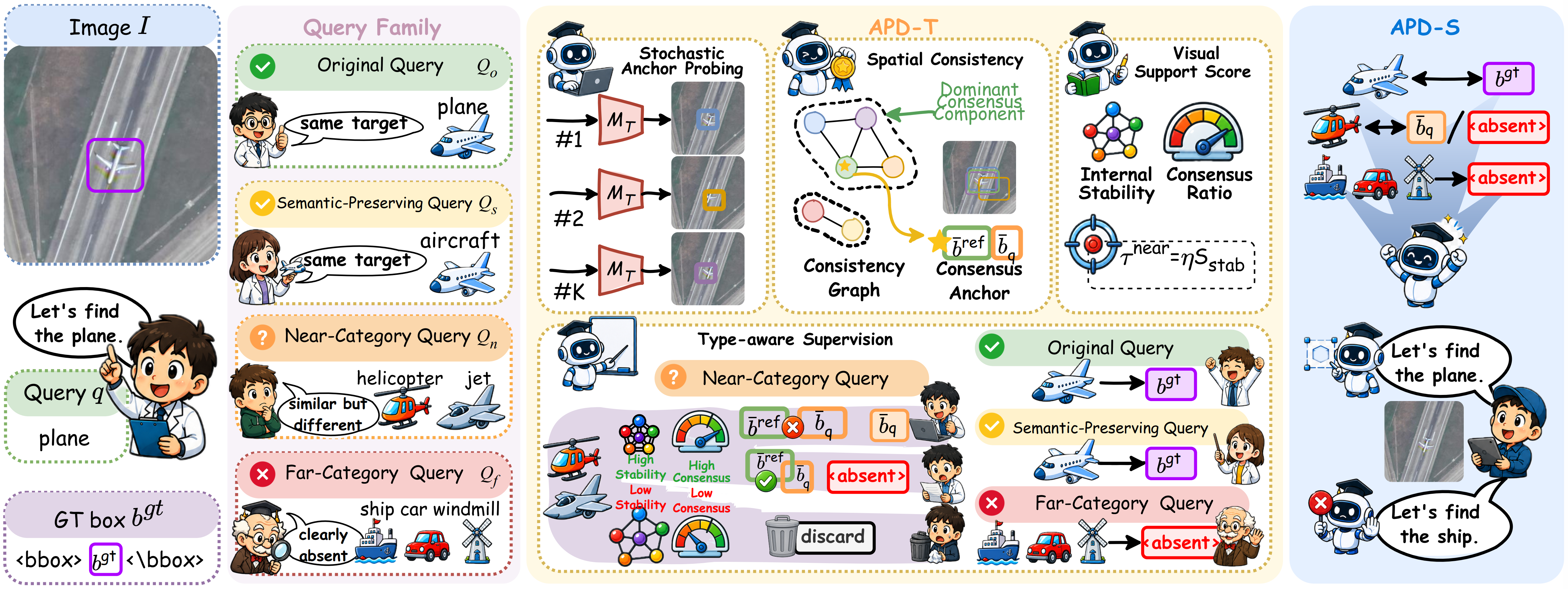}
\caption{AnchorProbe-Distill. APD-T converts type-aware stochastic probes into localization or rejection targets, which APD-S distills for single-pass open-set grounding.}
\label{fig:apd_framework}
\end{figure*}

Figure~\ref{fig:apd_framework} summarizes the pipeline: APD-T probes a type-aware query family, interprets Consensus Anchors relative to the Original Query, and generates localization or rejection targets; APD-S distills these targets so deployment requires only one image--query pass.

\subsection{Open-set Grounding as Visual Support Learning}

We denote the original positive training set as
\begin{equation}
\mathcal{D}=\{(I_i,q_i,b_i^{gt},c_i)\}_{i=1}^{N},
\end{equation}
where the $i$th training instance comprises a remote-sensing image $I_i\in\mathcal{I}$, a natural-language query $q_i\in\mathcal{Q}$, its annotated target box $b_i^{gt}\in\mathcal{B}$, and optional category metadata $c_i\in\mathcal{C}$. Standard grounding assumes that every $q_i$ is supported by its paired image $I_i$. In contrast, open-set grounding jointly predicts a binary grounding decision, a conditional localization, and a Support Confidence:
\begin{equation}
F_{\theta}(I,q)\rightarrow(\hat{y},\hat{b},\hat{C}),
\end{equation}
where $\hat{y}\in\{\bboxmath,\absentmath\}$ indicates whether the query is grounded or rejected, $\hat{b}\in\mathcal{B}\cup\{\varnothing\}$ is the predicted box list with $\hat{b}=\varnothing$ when $\hat{y}=\absentmath$, and $\hat{C}\in[0,1]$ quantifies support for the grounding-or-rejection decision. A grounded response follows \bboxformat{}, whereas an unsupported query produces \absentresp{} without a box.

APD does not require manually annotated negative queries. Instead, it constructs online supervision by perturbing each positive query into a typed family:
\begin{equation}
\mathcal{Q}(q_i)=
\mathcal{Q}_{o}\cup\mathcal{Q}_{s}\cup\mathcal{Q}_{n}\cup\mathcal{Q}_{f},
\quad
\mathcal{Q}_{o}=\{q_i\}.
\end{equation}
The four query sets encode different assumptions about Visual Support. $\mathcal{Q}_{o}$ contains the Original Query, queries in $\mathcal{Q}_{s}$ are Semantic-Preserving Queries that preserve the target identity, queries in $\mathcal{Q}_{n}$ are Near-Category Ambiguous Queries that describe visually or semantically adjacent categories, and queries in $\mathcal{Q}_{f}$ are Far-Category Counterfactual Queries that describe clearly different categories intended to be unsupported. A language model generates the query texts, which may be cached because their generation is independent of student updates; APD-T's support estimation and label assignment remain online during distillation. Before routing, category-family checks discard cross-family semantic queries and same-family near or far queries; unresolved language is retained conservatively. This decomposition is central: APD does not equate linguistic difference with negative supervision.

\subsection{Anchor Probing for Spatial Evidence}

APD-T estimates Visual Support from the stability of a query-conditioned spatial anchor under repeated teacher predictions. Given a fixed teacher grounding model $M_T$ and an image--query pair $(I,q)\in\mathcal{I}\times\mathcal{Q}$, APD-T performs $K$ stochastic teacher calls and obtains the candidate-box set
\begin{equation}
\mathcal{P}(I,q)=\{b_1,\ldots,b_{N_q}\}, \quad N_q\le K.
\end{equation}
The probing budget $K$ bounds the number of teacher calls, and $N_q$ denotes the number of valid boxes retained after response parsing. APD-T verifies the valid-box count before consensus estimation and treats the retained geometry as samples from a query-conditioned localization distribution. A visually supported query should repeatedly induce boxes around the same spatial anchor, whereas an unsupported query tends to produce scattered or unstable boxes.

To measure this concentration, APD-T constructs a CIoU consistency graph $G_q=(V_q,E_q)$. Each node corresponds to a sampled candidate box. Two boxes are connected when their geometric agreement exceeds a clustering threshold:
\begin{equation}
(u,v)\in E_q
\quad\Longleftrightarrow\quad
\operatorname{CIoU}(b_u,b_v)>\tau_{cluster}.
\end{equation}
The vertex set $V_q$ contains the sampled boxes, the edge set $E_q$ represents pairwise geometric consistency, and $\tau_{cluster}$ controls whether two boxes belong to the same consensus structure. CIoU incorporates overlap, center displacement, and aspect-ratio mismatch \cite{zheng2020distance}, enabling comparisons between spatially close predictions with different geometry. Each connected component of $G_q$ defines a Consensus Component. APD-T selects the largest component as the Dominant Consensus Component $\mathcal{C}_q$ and computes its coordinate-wise median as the Consensus Anchor $\bar{b}_q$:
\begin{equation}
\bar{b}_q=\operatorname{Median}\{b\mid b\in\mathcal{C}_q\}.
\end{equation}

The support estimate combines two complementary signals. Because CIoU lies in $[-1,1]$, we first define its normalized similarity as $s(b_m,b_l)=(\operatorname{CIoU}(b_m,b_l)+1)/2$. Internal Stability measures average pairwise agreement over all valid predictions:
\begin{equation}
S_{stab}(q)=
\frac{2}{N_q(N_q-1)}
\sum_{1\le m<l\le N_q}s(b_m,b_l).
\end{equation}
The summation covers all unordered pairs of valid boxes. This global statistic penalizes multimodal or scattered localization even when a compact Consensus Component can still be extracted. APD-T requires at least two valid boxes before evaluating this estimate.
The Consensus Ratio measures how much of the teacher distribution belongs to the Dominant Consensus Component:
\begin{equation}
R_{cons}(q)=\frac{|\mathcal{C}_q|}{N_q}.
\end{equation}
Thus, $S_{stab}(q)$ is the Internal Stability, measuring the global geometric concentration of valid predictions, while $R_{cons}(q)$ is the Consensus Ratio, measuring the share captured by the Dominant Consensus Component. Invalid-response frequency is handled by the separate Minimum Valid-Box Count rule rather than folded into either statistic.
The resulting Support Confidence is
\begin{equation}
C(q)=\frac{1}{2}S_{stab}(q)+\frac{1}{2}R_{cons}(q).
\end{equation}
The Support Confidence $C(q)$ assigns equal weight to anchor tightness and anchor frequency. A high value indicates that the teacher repeatedly returns a geometrically coherent anchor; a low value indicates insufficient Visual Support or unstable grounding behavior. APD attaches $C(q)$ to teacher-generated supervision but calibrates dynamic acceptance directly with Internal Stability $S_{stab}$. This separation isolates candidate geometry from the fraction of probes assigned to the Dominant Consensus Component.

\subsection{Type-aware Supervision Inference}

The main design choice in APD is that the four query types are not converted into supervision by the same rule. The Original Query and accepted Semantic-Preserving Queries keep the target identity and therefore always provide positive \bboxresp{} supervision. Far-Category Counterfactual Queries are used as explicit \absentresp{} supervision under the generated semantic assumption. Only Near-Category Ambiguous Queries require Type-Aware Supervision Inference, because they are intentionally ambiguous: a Near-Category Ambiguous Query may refer to a real unannotated object in the image, or it may be an unsupported category that a closed-set teacher collapses onto the original target. APD therefore uses the Original Query as a Reference Anchor and applies conservative near-query rules instead of treating all generated near queries as negatives.

For the Original Query $q_i\in\mathcal{Q}_{o}$, APD preserves the human annotation:
\begin{equation}
(I_i,q_i,\bboxmath,b_i^{gt},C(q_i)).
\end{equation}
Probing the teacher on $q_i$ preserves the human annotation while estimating the instance-specific Internal Stability and Reference Anchor. Denote the Consensus Anchor obtained from repeated Original-Query probes by $\bar{b}_i^{ref}$. The Reference Internal Stability determines the dynamic threshold
\begin{equation}
\tau_i=\eta S_{stab}(q_i),
\end{equation}
where $\eta\geq 1$ is calibrated per VLM from paired Original and Far-Category Counterfactual stability statistics, then reused unchanged as the multiplicative threshold for Near-Category routing. The rule adapts to the Original Query's Internal Stability level while requiring a Near-Category Ambiguous Query Consensus Anchor to be more reliable than the Reference Anchor.

For Semantic-Preserving Queries $q\in\mathcal{Q}_{s}$, APD directly inherits the original annotation:
\begin{equation}
(I_i,q,\bboxmath,b_i^{gt},C(q_i)).
\end{equation}
These queries change wording but not referent identity.

For each Near-Category Ambiguous Query $q\in\mathcal{Q}_{n}$, APD probes the teacher and estimates $(\bar{b}_q,S_{stab}(q),R_{cons}(q),C(q),N_q)$. Define
\(\mathcal{R}_{n}(q)\) as the Near-Query Reliability Test:
$S_{stab}(q)>\tau_i$, $R_{cons}(q)\geq\rho_{\min}$, and $N_q\geq N_{\min}$. A Near-Category Ambiguous Query is promoted to a new positive only when this reliability test is satisfied and the Consensus Anchor is spatially separated from the Reference Anchor:
\begin{equation}
\mathcal{R}_{n}(q)
\ \wedge\
\operatorname{IoU}(\bar{b}_q,\bar{b}_i^{ref})\leq\tau_{dis}.
\end{equation}
The Near-Query Reliability Test requires the near-query Consensus Anchor to have higher Internal Stability than the reference-calibrated threshold and to have enough probes agree on the Dominant Consensus Component. The IoU condition then requires the Consensus Anchor to be spatially separated from the Reference Anchor. APD uses the near-query Consensus Anchor as supervision:
\begin{equation}
(I_i,q,\bboxmath,\bar{b}_q,C(q)).
\end{equation}
Conversely, a reliable Near-Category Ambiguous Query that is semantically exclusive with the Original Query becomes a hard \absentresp{} example when its anchor collapses onto the Reference Anchor:
\begin{equation}
\mathcal{R}_{n}(q)
\ \wedge\
\operatorname{IoU}(\bar{b}_q,\bar{b}_i^{ref})\geq\tau_{ov}.
\end{equation}
APD then produces $(I_i,q,\absentmath,\varnothing,C(q))$. All other Near-Category Ambiguous Query cases are discarded rather than forced into absence supervision, including unstable anchors, low consensus ratio anchors, and anchors in the Overlap Dead Zone. This discard branch is important: failure to satisfy the positive pseudo-box rule means insufficient evidence, not necessarily target absence.

For Far-Category Counterfactual Queries $q\in\mathcal{Q}_{f}$, the semantic assumption is different. After removing any query that explicitly names the target family, APD uses the remaining distant-category queries as open-set negatives:
\begin{equation}
(I_i,q,\absentmath,\varnothing,0).
\end{equation}
This rule provides explicit rejection supervision during APD-S distillation. It is intentionally simple because Far-Category Counterfactual Queries are constructed to be semantically distant; ambiguous visual evidence is handled by Type-Aware Supervision Inference for Near-Category Ambiguous Queries rather than by relabeling every generated query through the same verifier.

\subsection{Online Distillation and Single-pass Inference}

For each online supervision instance, APD-S takes the image--query pair $(I_i,\tilde{q}_{ij})$ as input. Its target output is either \bboxformat{} for a grounded query or \absentresp{} for an unsupported query; box coordinates are normalized to the coordinate scale of the student VLM. APD-S follows the standard target-only autoregressive instruction-tuning objective \cite{liu2023llava}, while APD-T determines the online \bboxresp{}/\absentresp{} targets through Type-Aware Supervision Inference and its Consensus Anchor rules.

At inference time, APD-S no longer uses APD-T, query perturbations, stochastic sampling, or CIoU consensus. Given a test image and query, it directly predicts
\begin{equation}
F_{\theta}(I,q)\rightarrow(\hat{y},\hat{b},\hat{C}).
\end{equation}
If $\hat{y}=\bboxmath$, APD-S returns $\hat{b}$ in the \bboxresp{} response together with $\hat{C}$ as its predicted Support Confidence; if $\hat{y}=\absentmath$, it emits \absentresp{} and uses $\hat{C}$ as the confidence of the rejection decision. APD therefore shifts both the Reference Anchor information and the computation of Visual Support verification from deployment to training. APD-T requires $K$ probes for a supported query and, in paired Far-Category Counterfactual Query verification, $K$ probes for both the Reference Anchor and candidate queries. APD-S retains the single-query, single-pass structure of a standard grounding model while additionally exposing a Support Confidence estimate.

\begin{algorithm}[t]\footnotesize
\caption{AnchorProbe-Distill Training}
\label{alg:anchorprobe_distill_online}
\begin{algorithmic}[1]
\REQUIRE Positive dataset $\mathcal{D}$, teacher $M_T$, student $F_{\theta}$, probes $K$
\ENSURE Distilled open-set grounding model $F_{\theta}$
\FOR{each training iteration}
\STATE Sample a mini-batch $\mathcal{M}\subset\mathcal{D}$
\STATE Initialize supervision set $\mathcal{S}\leftarrow\emptyset$
\FOR{each $(I_i,q_i,b_i^{gt},c_i)\in\mathcal{M}$}
\STATE Set $\mathcal{Q}_{o}=\{q_i\}$; generate $\mathcal{Q}_{s}$, $\mathcal{Q}_{n}$, and $\mathcal{Q}_{f}$
\STATE Remove category-inconsistent $\mathcal{Q}_{s}$ and same-family $\mathcal{Q}_{n}$ queries
\STATE Probe $M_T(I_i,q_i)$ and compute $\bar{b}_i^{ref}$, $S_{stab}(q_i)$, $C(q_i)$, and $\tau_i=\eta S_{stab}(q_i)$
\STATE Add positives from $\mathcal{Q}_{o}$ and $\mathcal{Q}_{s}$ to $\mathcal{S}$
\FOR{each $q\in\mathcal{Q}_{n}$}
\STATE Probe $M_T(I_i,q)$ and estimate $(\bar{b}_q,S_{stab}(q),R_{cons}(q),C(q),N_q)$
\IF{$N_q\geq N_{\min}$, $S_{stab}(q)>\tau_i$, $R_{cons}(q)\geq\rho_{\min}$, and $\operatorname{IoU}(\bar{b}_q,\bar{b}_i^{ref})\leq\tau_{dis}$}
\STATE Add $(I_i,q,\bboxmath,\bar{b}_q,C(q))$ to $\mathcal{S}$
\ELSIF{$N_q\geq N_{\min}$, $S_{stab}(q)>\tau_i$, $R_{cons}(q)\geq\rho_{\min}$, and $\operatorname{IoU}(\bar{b}_q,\bar{b}_i^{ref})\geq\tau_{ov}$}
\STATE Add $(I_i,q,\absentmath,\varnothing,C(q))$ to $\mathcal{S}$
\ENDIF
\ENDFOR
\FOR{each $q\in\mathcal{Q}_{f}$}
\STATE Add $(I_i,q,\absentmath,\varnothing,0)$ to $\mathcal{S}$
\ENDFOR
\ENDFOR
\STATE Update $F_{\theta}$ with the online supervision in $\mathcal{S}$
\ENDFOR
\RETURN $F_{\theta}$
\end{algorithmic}
\end{algorithm}

\section{Experiments}

\subsection{Experimental Setup}

The evaluation uses the RefGeo test set and its paired RefGeo Far-Category test set, each containing 37,763 samples from AVVG~\cite{zhou2024geoground}, DIOR-RSVG~\cite{zhan2023rsvg}, GeoChat~\cite{kuckreja2023geochat}, RRSISD~\cite{liu2024rmsin}, RSVG~\cite{sun2022visualgrounding}, and VRSBench~\cite{li2024vrsbench}. The RefGeo test set measures supported-query localization, whereas the RefGeo Far-Category test set replaces the target phrase with an unsupported query on the same image. Under the same strict verifier, the comparison includes Qwen3-VL-2B, LLaVA-7B, their RefGeo-fine-tuned variants, GeoGround, Geo-R1-3B~\cite{zhang2026geor1}, RemoteReasoner~\cite{yao2025remotereasoner}, GeoPixel~\cite{shabbir2025geopixel}, and Text4Seg~\cite{lan2025text4seg}. Published results obtained under non-unified data and evaluation protocols are separated in Appendix~\ref{app:cross_protocol_results}.

Acc@0.5 is the percentage of RefGeo predictions with IoU$>0.5$. Reject Ratio (RR) counts responses that are empty or unambiguously state that the target is absent; nonempty malformed outputs are failures. Unless otherwise stated, APD uses $K=10$, $\tau_{cluster}=0.5$, $N_{min}=2$, $\rho_{min}=0.4$, and the VLM-specific $\eta$ values in Appendix~\ref{app:experimental_details}. Other implementation details are also provided there.

\begin{figure}[!t]
\centering
\includegraphics[width=\columnwidth]{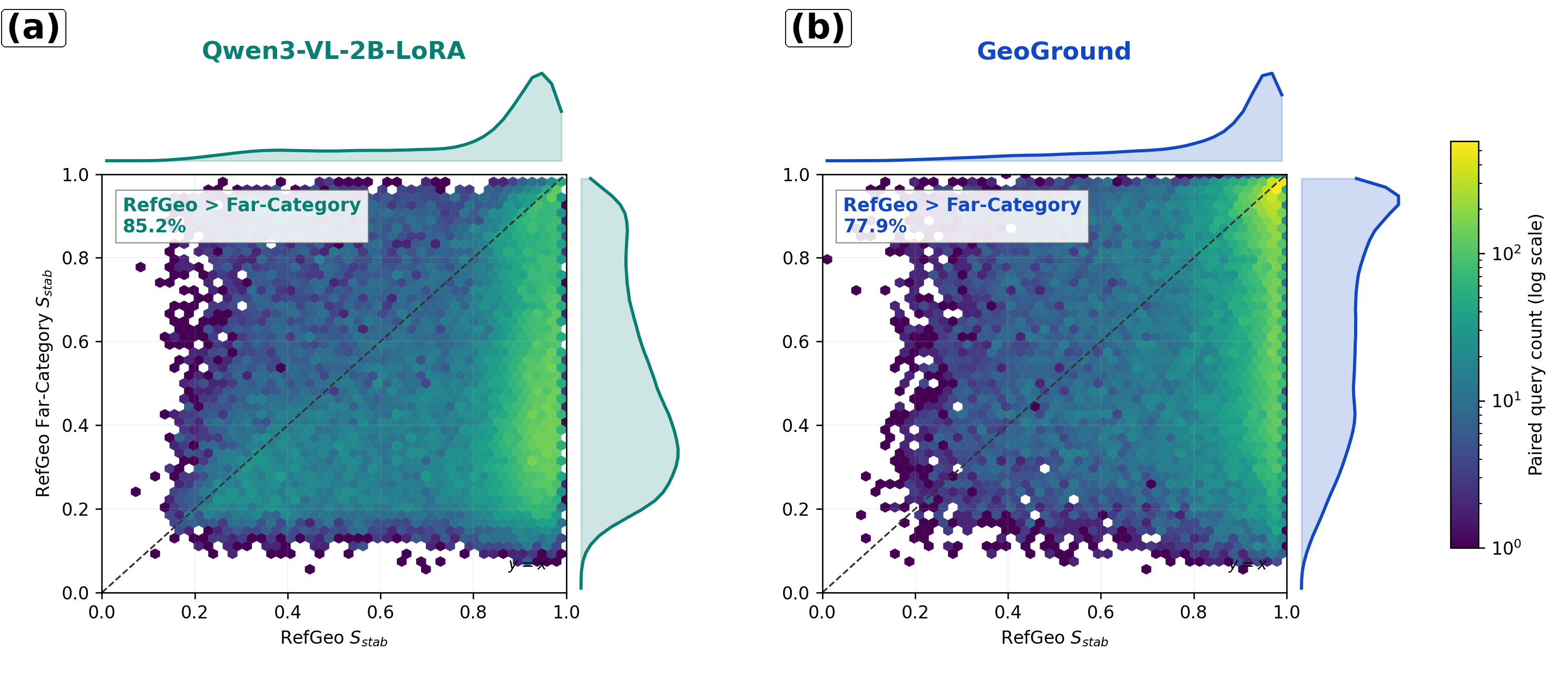}
\caption{Internal Stability of paired supported and unsupported queries for (a) Qwen3-VL-2B-FT and (b) GeoGround. Color shows pair count on a shared log scale; zero scores are omitted.}
\label{fig:positive_negative_stability}
\end{figure}

\subsection{Open-Set Grounding Performance}

\begin{table*}[t]
\centering
\footnotesize
\setlength{\tabcolsep}{3.0pt}
\renewcommand{\arraystretch}{1.18}

\begin{tabular}{@{}M*{7}{NN}@{}}
\toprule
\multirow{2}{*}{Method}
& \multicolumn{2}{c}{AVVG}
& \multicolumn{2}{c}{DIOR-RSVG}
& \multicolumn{2}{c}{GeoChat}
& \multicolumn{2}{c}{RRSISD}
& \multicolumn{2}{c}{RSVG}
& \multicolumn{2}{c}{VRSBench}
& \multicolumn{2}{c}{Overall} \\
\cmidrule(lr){2-3}\cmidrule(lr){4-5}\cmidrule(lr){6-7}\cmidrule(lr){8-9}\cmidrule(lr){10-11}\cmidrule(lr){12-13}\cmidrule(lr){14-15}
& A & RR & A & RR & A & RR & A & RR & A & RR & A & RR
& A & RR \\
\midrule

\multicolumn{15}{@{}l}{\textbf{\textit{General models}}} \\
Qwen3-VL-2B
& 14.90 & 44.31 & 49.23 & 22.27 & 44.62 & 19.68 & 51.54 & 21.32 & 19.72 & 16.06 & 45.82 & 10.99
& 38.65 & 23.24 \\
LLaVA-7B
& 1.07 & 60.72 & 6.40 & 52.45 & 6.37 & 57.22 & 8.04 & 54.22 & 0.16 & 44.82 & 5.88 & 53.86
& 4.90 & 55.36 \\

\addlinespace[1pt]
\multicolumn{15}{@{}l}{\textbf{\textit{Fine-tuned models}}} \\
Qwen3-VL-2B$^\S$
& 27.19 & 0.20 & 79.98 & 0.00 & 70.90 & 0.00 & 83.55 & 0.00 & 42.30 & 0.00 & 70.92 & 0.00
& 62.48 & 0.05 \\
LLaVA-7B$^\S$
& 0.38 & 0.00 & 22.86 & 0.00 & 12.99 & 0.00 & 26.68 & 0.00 & 0.81 & 0.00 & 14.17 & 0.02
& 13.10 & 0.01 \\

\addlinespace[1pt]
\midrule
\multicolumn{15}{@{}l}{\textbf{\textit{Specialist models}}} \\
GeoGround
& 17.78 & 0.00 & 77.34 & 0.00 & 66.75 & 0.00 & 91.28 & 0.00 & 26.81 & 0.00 & 65.42 & 0.00
& 57.60 & 0.00 \\
Geo-R1-3B
& 13.33 & 0.42 & 41.70 & 2.38 & 36.50 & 3.03 & 45.35 & 2.53 & 12.88 & 1.63 & 40.69 & 1.59
& 33.38 & 1.73 \\
RemoteReasoner
& 13.71 & 0.25 & 42.68 & 0.09 & 41.07 & 0.04 & 44.02 & 0.03 & 9.37 & 0.00 & 36.22 & 0.11
& 32.55 & 0.12 \\
GeoPixel
& 5.91 & 0.00 & 68.89 & 0.00 & 43.18 & 0.00 & 71.20 & 0.00 & 15.65 & 0.00 & 51.88 & 0.00
& 43.73 & 0.00 \\
Text4Seg
& 3.22 & 0.00 & 27.58 & 0.00 & 23.85 & 0.00 & 27.30 & 0.00 & 0.49 & 0.00 & 25.41 & 0.02
& 19.73 & 0.01 \\
\addlinespace[1pt]
\midrule
\multicolumn{15}{@{}l}{\textbf{\textit{Proposed APD variants}}} \\
GeoGround$^{\dagger}$
& 18.76 & 65.12 & 78.08 & 87.66 & 70.45 & 72.67 & \textbf{92.73} & 91.34 & 27.47 & 77.59 & 66.22 & 91.46
& 58.84 & 81.68 \\
Qwen3-VL-2B$^{\dagger}$
& 30.14 & 71.90 & \textbf{84.92} & 92.85 & \textbf{76.54} & 90.64 & 88.05 & 95.22 & \textbf{54.44} & 85.82 & \textbf{77.76} & 93.46
& \textbf{67.85} & 87.78 \\
Qwen3-VL-2B$^{\ddagger}$ & \textbf{31.07} & \textbf{99.61} & 83.08 & \textbf{95.89} & 73.85 & \textbf{98.63} & 86.14 & \textbf{96.64} & 53.14 & \textbf{93.89} & 75.07 & \textbf{98.04} & 66.30 & \textbf{97.81} \\

\bottomrule
\end{tabular}

\vspace{2pt}
\begin{minipage}{0.98\textwidth}
\footnotesize
$^\S$ fine-tuned; $^\dagger$ APD-T; $^{\ddagger}$ APD-S.
\end{minipage}
\caption{Open-set grounding on RefGeo (\%). \textbf{A}: Acc@0.5; \textbf{RR}: Reject Ratio on the paired Far-Category set. \textbf{Bold} marks the best reported value per column.}
\label{tab:main_results}
\end{table*}

Table~\ref{tab:main_results} reports source-wise and sample-weighted pooled performance. Beyond localization, APD addresses whether a spatial prediction is justified by the image. Geo-R1-3B, RemoteReasoner, GeoPixel, and Text4Seg achieve at most 1.73\% RR despite their localization ability, demonstrating that accurate grounding alone does not provide reliable rejection. APD-T improves Qwen3-VL-2B from 62.48/0.05 to 67.85/87.78 in Overall Acc@0.5/RR and GeoGround from 57.60/0.00 to 58.84/81.68, with consistent rejection gains across all six sources. The Qwen3-VL-2B APD-S further achieves 66.30 Acc@0.5 and 97.81 RR, retaining the localization gain while attaining the strongest rejection result. Appendix Figures~\ref{fig:external_supported_outputs} and~\ref{fig:external_absent_outputs} compare the models on common image--query inputs: the evaluated VLMs localize supported targets but return boxes for an absent target, whereas the APD variants correctly reject the unsupported query.

\subsection{Visual Support Stability Analysis}

The stability analysis tests the hypothesis introduced in Figure~\ref{fig:intro_sampling}: repeated probes of a visually supported query should concentrate around a stable target, whereas probes of an unsupported query should be more dispersed. Figure~\ref{fig:positive_negative_stability} evaluates this hypothesis at test-set scale by pairing each Original Query with its Far-Category Counterfactual Query on the same image. The dominant density below the diagonal indicates that $S_{stab}(q_o)>S_{stab}(q_f)$ for a large proportion of pairs, demonstrating that supported targets generally induce more stable localization than unsupported targets. The repeated-prediction overlays in Figure~\ref{fig:intro_sampling} and Appendix Figure~\ref{fig:cross_model_qualitative} provide complementary qualitative evidence. This observation directly supports geometric stability as the verification signal exploited by APD-T.

\subsection{Grouped APD-T Ablations}

The probe-budget ablation examines APD-T sensitivity through offline post-processing, requiring neither new inference nor APD-S updates. Table~\ref{tab:ablation_k} demonstrates that the default budget is close to saturation for most VLMs. Larger budgets improve LLaVA localization but leave RR nearly unchanged; the default exploits 60\% fewer probes than $K=25$ while retaining comparable performance.

\begin{table}[t]
\centering
\footnotesize
\setlength{\tabcolsep}{3.2pt}
\renewcommand{\arraystretch}{0.94}
\begin{tabular}{llcccc}
\toprule
VLM & Metric & $K{=}5$ & $K{=}10$ & $K{=}25$ & $K{=}50$ \\
\midrule
\multirow{2}{*}{GeoGround}
& Acc@0.5 & 58.66 & 58.84 & 58.82 & \textbf{58.87} \\
& RR & 80.37 & 81.68 & \textbf{82.17} & 82.05 \\
\addlinespace[1pt]
\multirow{2}{*}{LLaVA-7B}
& Acc@0.5 & 5.66 & 6.29 & 7.75 & \textbf{8.69} \\
& RR & 99.05 & \textbf{99.38} & 99.36 & 99.32 \\
\multirow{2}{*}{LLaVA-7B$^{\S}$}
& Acc@0.5 & 14.87 & 15.37 & 16.38 & \textbf{16.75} \\
& RR & 96.73 & \textbf{98.01} & 97.73 & 97.31 \\
\addlinespace[1pt]
\multirow{2}{*}{Qwen3-VL-2B}
& Acc@0.5 & 40.56 & 40.90 & 41.11 & \textbf{41.25} \\
& RR & 87.04 & \textbf{87.79} & 87.68 & 87.72 \\
\multirow{2}{*}{Qwen3-VL-2B$^{\S}$}
& Acc@0.5 & 67.69 & 67.85 & 67.98 & \textbf{68.10} \\
& RR & 86.65 & 87.78 & 88.36 & \textbf{88.49} \\
\bottomrule
\end{tabular}
\caption{APD-T probe-budget ablation on paired RefGeo/Far-Category sets (\%).
\textbf{Bold} marks the best value per VLM and metric; $^\S$ denotes
fine-tuned.}
\label{tab:ablation_k}
\end{table}

The cluster-threshold ablation varies $\tau_{cluster}$ from 0.1 to 0.9 using the cached default-budget predictions. Table~\ref{tab:ablation_tau_cluster} reports Consensus Anchor Acc@0.5 and mean $R_{cons}$. Qwen3-VL-2B and GeoGround remain stable through 0.5, whereas stricter thresholds fragment the consensus graph and progressively reduce $R_{cons}$. This transition supports $\tau_{cluster}=0.5$ as a stable operating point.

\begin{table}[t]
\centering
\footnotesize
\setlength{\tabcolsep}{2.0pt}
\begin{tabular}{c*{3}{cc}}
\toprule
& \multicolumn{2}{c}{Qwen3-VL-2B$^{\S}$} & \multicolumn{2}{c}{GeoGround} & \multicolumn{2}{c}{LLaVA-7B$^{\S}$} \\
$\tau_{cluster}$ & Acc@0.5 & $R_{cons}$ & Acc@0.5 & $R_{cons}$ & Acc@0.5 & $R_{cons}$ \\
\midrule
0.1 & 68.32 & 0.872 & 58.93 & 0.931 & 16.71 & 0.626 \\
0.2 & 68.38 & 0.866 & 58.96 & 0.924 & 16.75 & 0.590 \\
0.3 & 68.40 & 0.862 & 58.98 & 0.913 & \textbf{16.89} & 0.548 \\
0.4 & 68.41 & 0.857 & \textbf{59.00} & 0.901 & 16.85 & 0.499 \\
0.5 & \textbf{68.50} & 0.849 & \textbf{59.00} & 0.882 & 16.63 & 0.439 \\
0.6 & 68.39 & 0.838 & 58.98 & 0.860 & 16.40 & 0.366 \\
0.7 & 68.27 & 0.811 & 58.90 & 0.826 & 15.89 & 0.278 \\
0.8 & 67.99 & 0.745 & 58.75 & 0.760 & 14.94 & 0.183 \\
0.9 & 67.40 & 0.566 & 58.45 & 0.623 & 13.60 & 0.114 \\
\bottomrule
\end{tabular}
\caption{Consensus Anchor's performance versus $\tau_{cluster}$.}
\label{tab:ablation_tau_cluster}
\end{table}

\begin{figure}[t]
\centering
\includegraphics[width=\columnwidth]{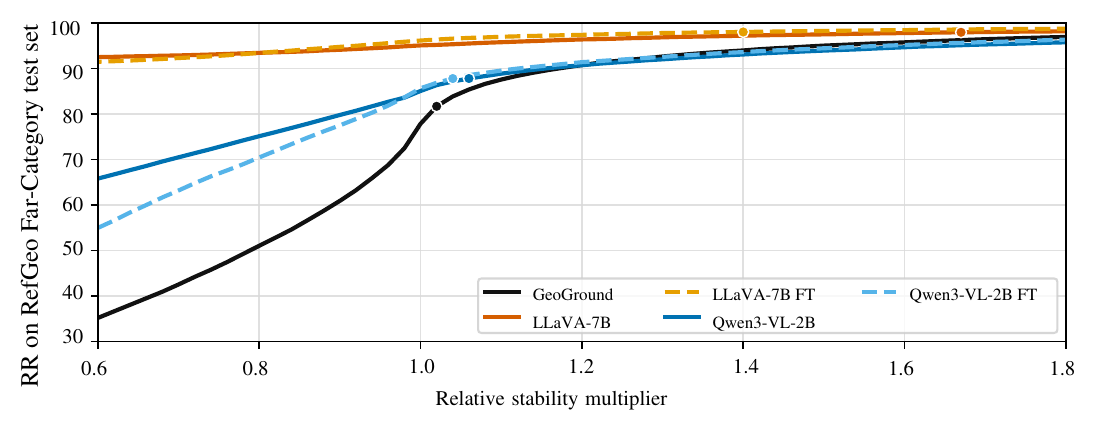}
\caption{APD-T RR versus $\eta$. Dots mark the largest value with at least 80\% score feasibility. The selected per-VLM value is reused for Near-Category routing.}
\label{fig:eta_ablation}
\end{figure}

The final APD-T ablation evaluates $\eta\in[0.6, 1.8]$ under the default protocol. Selecting $\eta$ by RR alone would favor an overly strict endpoint. Because $S_{stab}\leq 1$, a query can clear the relative stability gate only when $\eta S_0<1$. Accordingly, the score-feasibility ceiling is defined as
$F_M(\eta)=N_M^{-1}\sum_i\mathbf{1}[\eta S_{0,i}<1]$, and the selected operating point is the largest swept $\eta$ satisfying $F_M(\eta)\geq0.80$. The operating points in Figure~\ref{fig:eta_ablation} favor stronger rejection without making the gate unattainable for most reference scores.

\subsection{APD-S Supervision and Routing}

\begin{figure}[t]
\centering
\includegraphics[width=\columnwidth]{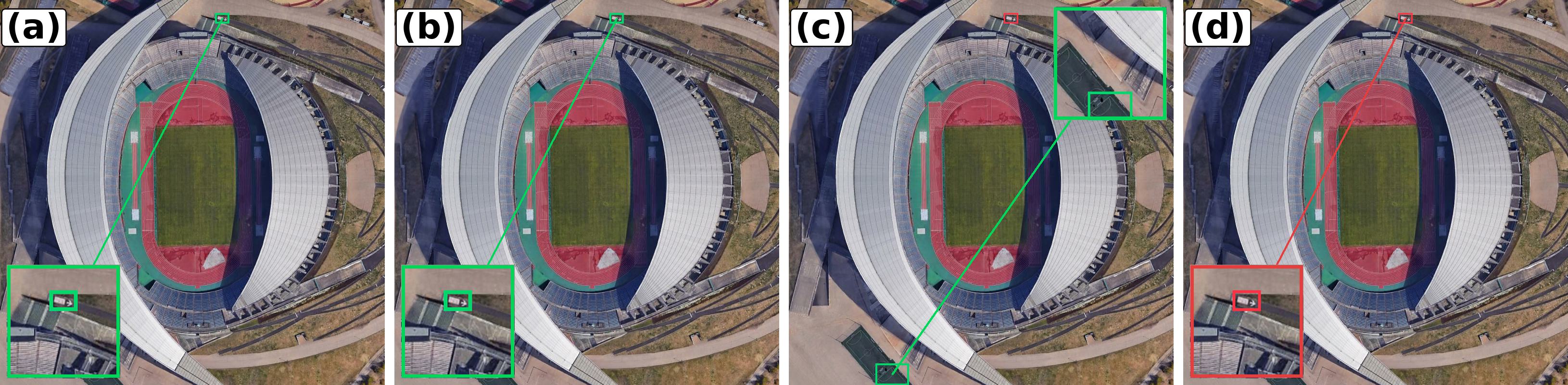}
\caption{APD-T supervision: (a) Original, ``The tiny vehicle on the top''; (b) Semantic-Preserving, ``a small vehicle on the top of the stadium''; (c) accepted Near-Category, ``a basketball court in the bottom''; (d) Far-Category, ``a storage tank in the middle of the field.'' Panels (a)--(c) yield box targets; (d) yields \absentresp{}. Red/green mark the Reference/APD target; insets enlarge target regions.}
\label{fig:online_supervision_queries}
\end{figure}

\begin{figure}[t]
\centering
\includegraphics[width=0.8\columnwidth]{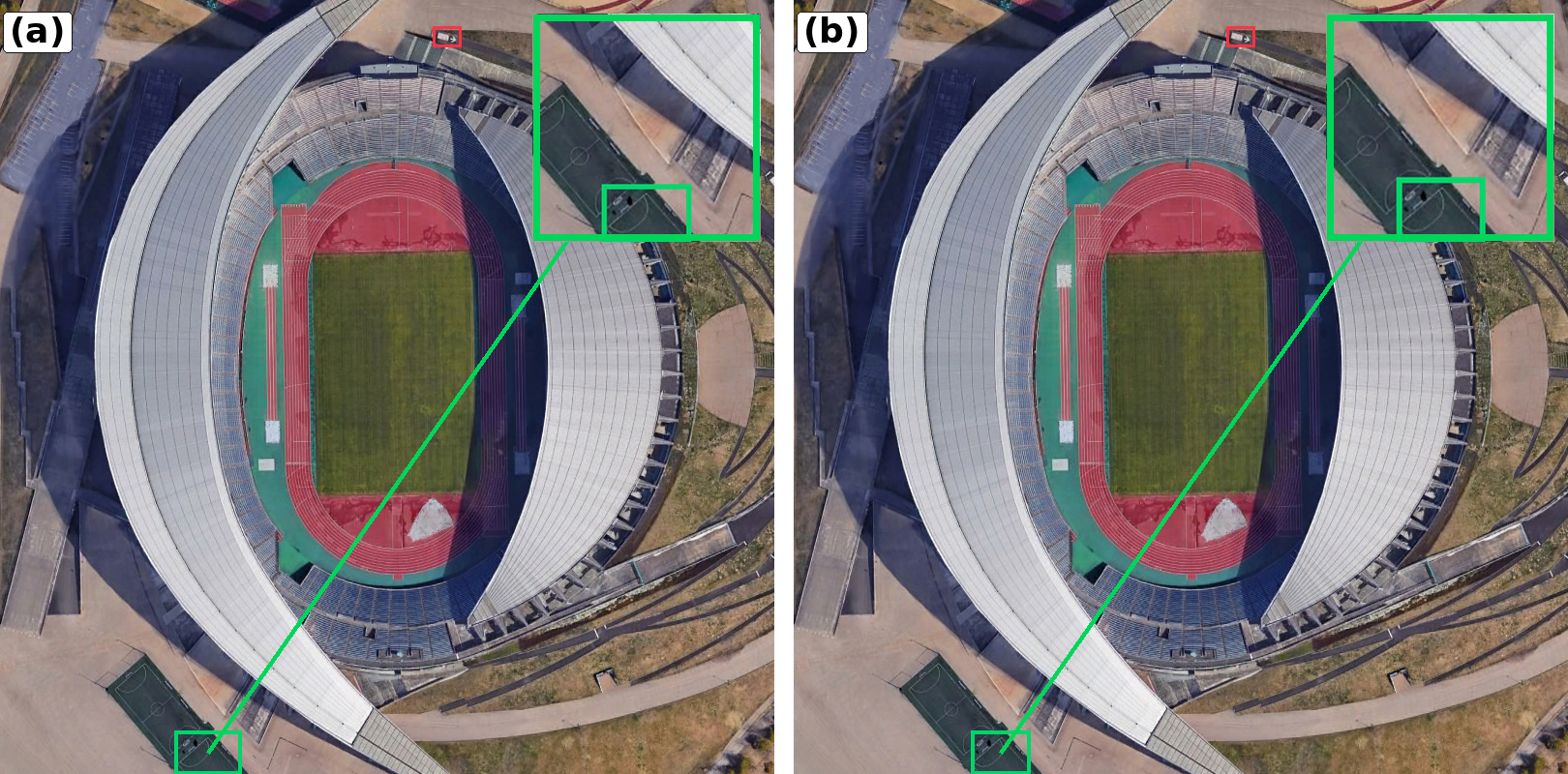}
\caption{Near-Category routing ($S_0=0.836$, $\eta=1.04$, $\tau=0.869$). (a) ``a basketball court in the bottom'': $S_{stab}=0.929$, $R_{cons}=1.0$, disjoint positive. (b) ``a tennis court in the bottom'': $S_{stab}=0.841$, $R_{cons}=1.0$, below threshold and unused. Red/green mark the Reference/candidate anchor.}
\label{fig:online_supervision_near_routing}
\end{figure}

Figures~\ref{fig:online_supervision_queries} and \ref{fig:online_supervision_near_routing} visualize which online queries are accepted by APD-T and which targets are delivered to APD-S. Figure~\ref{fig:online_supervision_queries} makes the routing concrete: the Original and Semantic-Preserving Queries retain the annotated target, the Near-Category Ambiguous Query supplies a spatially disjoint positive anchor, and the Far-Category Counterfactual Query supplies \absentresp{} supervision. Every displayed query is delivered to the applicable student. Figure~\ref{fig:online_supervision_near_routing} contrasts the two Near-Category decisions against the same Original Query: $S_0=0.8356$ and $\eta=1.04$ give $\tau=0.8690$. The filtered candidate in (b) contributes no APD-S loss.

\subsection{Qualitative Semantic Alignment}

\begin{figure}[!t]
\centering
\includegraphics[width=\columnwidth]{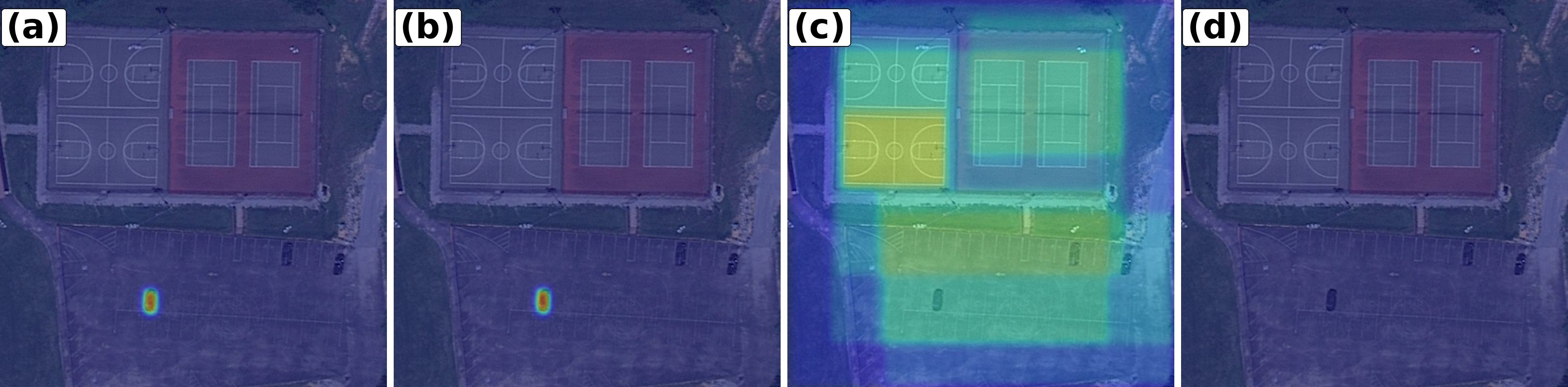}
\caption{Query-conditioned box density before/after APD-S: (a,b) supported ``1 gray vehicle at the bottom of the image''; (c,d) absent ``Locate the golf course''; (a,c) before and (b,d) after training. Red/blue indicate high/low density.}
\label{fig:apds_pre_post_heatmaps}
\end{figure}

Beyond output-level accuracy and rejection, the spatial response should remain aligned with the queried semantics. Figure~\ref{fig:apds_pre_post_heatmaps} illustrates how APD changes the relation between query semantics and spatial activation. The supported vehicle query remains localized, whereas the absent golf-course query no longer attaches to visually salient court and parking textures. The contrast indicates that APD-S conditions its response more strongly on image--query support rather than scene similarity alone.

\subsection{APD-S Supervision Ablation}

\begin{table}[!t]
\centering
\small
\begin{tabular}{lcc}
\toprule
VLM / training query sets & Acc@0.5 & RR \\
\midrule
Qwen3-VL-2B$^\S$ & 62.48 & 0.05 \\
$\mathcal{Q}_{o}+\mathcal{Q}_{s}$ & 64.28 & 0.00 \\
$\mathcal{Q}_{o}+\mathcal{Q}_{s}+\mathcal{Q}_{n}$ & 65.64 & 2.23 \\
$\mathcal{Q}_{o}+\mathcal{Q}_{s}+\mathcal{Q}_{n}+\mathcal{Q}_{f}$ & \textbf{66.30} & \textbf{97.81} \\
\bottomrule
\end{tabular}
\caption{Performance under different APD-S supervision.}
\label{tab:ablation_perturbation}
\end{table}

To determine which query types drive localization and rejection, we incrementally add the supervision families routed by APD-T. In Table~\ref{tab:ablation_perturbation}, all variants share the same Qwen3-VL-2B initialization and APD-T stream, differing only in routed supervision. Table~\ref{tab:ablation_perturbation} isolates each query family's role. Semantic-Preserving Queries raise Acc@0.5 from 62.48 to 64.28 but leave RR at 0, consistent with preserving the original referent. Accepted Near-Category Queries add another 1.36 localization points while raising RR only to 2.23, indicating that conservative near-category routing primarily enriches positive supervision. Far-Category Counterfactual Queries add 0.66 localization points and increase RR by 95.58 points to 97.81. These trends attribute localization coverage to Semantic-Preserving and accepted Near-Category supervision, and the explicit absence signal to Far-Category supervision.

\section{Conclusion}
\label{sec:conclusion}

We present AnchorProbe-Distill (APD), which recasts single-target-or-absent remote-sensing grounding as Visual Support learning. APD-T converts repeated predictions into reference-calibrated, type-aware targets, and APD-S distills this verification capability for single-pass inference. On RefGeo, APD-S achieves 66.30\% Acc@0.5 and 97.81\% RR. Ablations attribute localization gains to Semantic-Preserving and accepted Near-Category Queries, and rejection to Far-Category Counterfactual Queries. Repeated-prediction stability thus provides scalable supervision without manually annotated negative pairs and is readily distillable for deployment, establishing rejection as a first-class objective for open-world grounding.

\bibliography{aaai2027}

\clearpage
\appendix

\section{Supplementary Material}

\section{Related Work}

\subsection{Remote-Sensing Vision--Language Models}

\begin{table*}[t]
\centering
\setlength{\tabcolsep}{2.0pt}
\begin{tabular}{lrrrrrrr}
\toprule
Method & AVVG & DIOR-RSVG & GeoChat & RRSISD & RSVG & VRSBench & Overall \\
\midrule
\multicolumn{8}{l}{\textit{Published, non-unified protocols}} \\
GeoChat~\cite{kuckreja2023geochat}
& 0.28 & 24.05 & 22.74 & \na & 2.04 & 11.52 & \na \\
EarthGPT~\cite{zhang2024earthgpt}
& \na & 76.65 & \na & \na & \na & \na & \na \\
SkySenseGPT~\cite{luo2024sky}
& 26.50 & 60.80 & \na & 69.00 & 39.50 & 63.50 & \na \\
GeoViS~\cite{zhang2026geovis}
& \na & 79.80 & 23.70 & \na & \na & 68.50 & \na \\
SkyMoE~\cite{liu2026skymoe}
& 36.50 & 68.60 & \na & 72.50 & 42.50 & 66.10 & \na \\
\midrule
\multicolumn{8}{l}{\textit{Ours, unified RefGeo protocol}} \\
GeoGround APD-T
& 18.76 & 78.08 & 70.45 & 92.73 & 27.47 & 66.22 & 58.84 \\
Qwen3-VL-2B APD-T
& 30.14 & 84.92 & 76.54 & 88.05 & 54.44 & 77.76 & 67.85 \\
Qwen3-VL-2B APD-S
& 31.07 & 83.08 & 73.85 & 86.14 & 53.14 & 75.07 & 66.30 \\
\bottomrule
\end{tabular}%
\caption{Localization under published and unified protocols (\%). Upper: paper-reported Acc@0.5/Pr@0.5; lower: our RefGeo evaluation. The blocks are not directly comparable.}
\label{tab:cross_protocol_results}
\end{table*}

Early remote-sensing vision--language research established visual question
answering and language-conditioned scene understanding as alternatives to
fixed-label recognition. RSVQA introduced large-scale question answering over
overhead imagery, while Prompt-RSVQA and subsequent surveys studied stronger
visual--linguistic interfaces and the broader role of language in Earth
observation \cite{lobry2020rsvqa,chappuis2022promptrsvqa,bashmal2023language}.
More recent foundation models scale this direction through contrastive
pretraining and large image--text corpora. RemoteCLIP targets transferable
remote-sensing representations, RS5M/GeoRSCLIP expands multimodal pretraining
data, and SkySense combines multiple sensing modalities for general Earth
observation interpretation
\cite{liu2024remoteclip,zhang2023rs5m,guo2024skysense}. Instruction-tuned
systems including LHRS-Bot, GeoChat, EarthGPT, SkySenseGPT, and SkyMoE further
support dialogue, reasoning, and fine-grained spatial descriptions
\cite{muhtar2024lhrsbot,kuckreja2023geochat,zhang2024earthgpt,luo2024sky,liu2026skymoe}.
These developments supply increasingly capable VLMs, but their standard
instruction objectives do not by themselves ensure that a spatial response
is supported by the image.

\subsection{Referring Grounding and Geospatial Reasoning}

Referring expression comprehension maps a free-form description to a spatial
region. Generalized REC extends the conventional single-referent setting,
and later work improves generalized grounding through hierarchical alignment
\cite{he2023grec,wang2025hierarchical}. RefDrone exposes the difficulty of
referring comprehension in drone imagery, while region-aware anchoring
improves the efficiency of grounding visual regions
\cite{sun2025refdrone,ouyang2025region}. In remote sensing, GeoGround
introduces RefGeo and unifies bounding boxes, oriented boxes, and masks;
GeoViS instead casts grounding as a geospatially rewarded visual search
\cite{zhou2024geoground,zhang2026geovis}. Geo-R1 applies reinforcement
fine-tuning to few-shot geospatial referring expressions, and
RemoteReasoner emphasizes a unified geospatial reasoning workflow
\cite{zhang2026geor1,yao2025remotereasoner}. These approaches mainly optimize
where to localize a described target. APD addresses the complementary
question of whether the image supports returning any location, and it can be
applied on top of different grounding VLMs rather than replacing their
localization architecture.

\section{Detailed Experimental Protocol}
\label{app:experimental_details}

\subsection{Implementation and Computing Environment}

Experiments are conducted on a server with four NVIDIA A800 80\,GB GPUs,
two Intel Xeon Platinum 8352V CPUs, and 1\,TB of system memory, running Ubuntu
22.04. The implementation uses Python 3.11, PyTorch 2.10, CUDA 12.8,
Transformers 5.2, and PEFT 0.18. APD-S is optimized with AdamW using a constant
learning rate of $2\times10^{-5}$ and zero weight decay. LoRA parameters are
updated with source batch size 1; the supervision microbatch size is 8 for
Qwen3-VL-2B and 2 for GeoGround. Each student is trained for one epoch, i.e.,
one complete online pass over the independently shuffled RefGeo training
stream. APD-T remains frozen throughout probing and supervision generation.

\subsection{Datasets and Pairing}

The paired test sets use identical source-image order and preserve source identifiers for APD-T calibration. Their source counts are AVVG (8,956), DIOR-RSVG (7,422), GeoChat (4,680), RRSISD (3,246), RSVG (1,227), and VRSBench (12,232). The Far-Category test set is used only for evaluation. Training and evaluation iterators are independently shuffled with seed 42.

\begin{figure}[!b]
\centering
\includegraphics[width=\columnwidth]{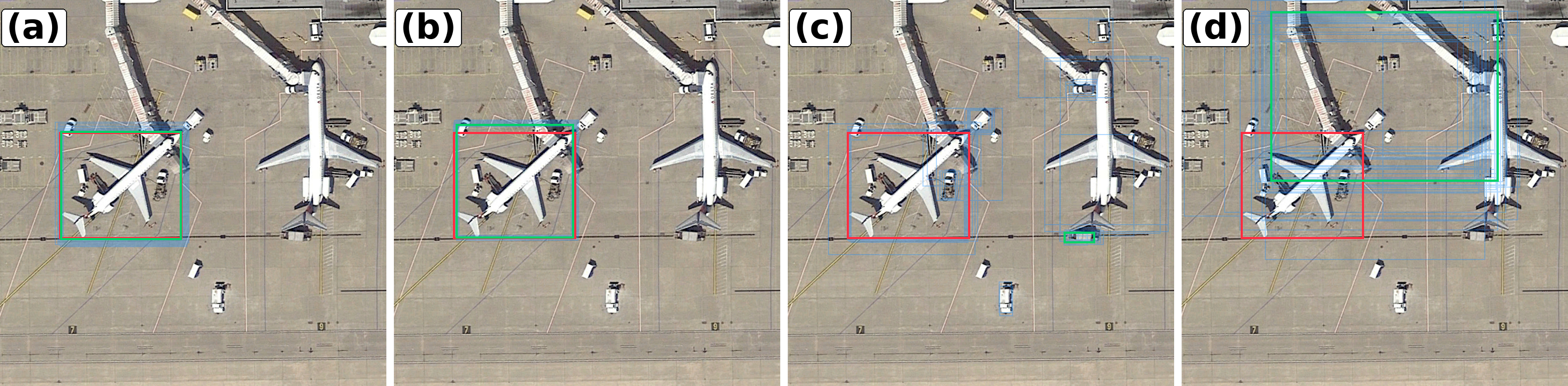}
\caption{Repeated predictions on one DIOR-RSVG image: (a,c) Qwen3-VL-2B-FT; (b,d) GeoGround; (a,b) airplane query; (c,d) dam query. Blue: 50 samples; red: Reference Anchor; green: first prediction.}
\label{fig:cross_model_qualitative}
\end{figure}

\subsection{Compared Methods}

We evaluate base/RefGeo-fine-tuned Qwen3-VL, LLaVA, GeoGround, and APD variants under a common evaluator. Geo-R1-3B-GRPO-REC-10shot, RemoteReasoner, GeoPixel, and Text4Seg also provide complete identifier-aligned predictions. Published results using non-unified protocols are retained only as contextual references below.

\subsection{Cross-Protocol Published Results}
\label{app:cross_protocol_results}

Table~\ref{tab:cross_protocol_results} separates published localization results
from results produced by our unified RefGeo evaluator. The upper block is
transcribed from the respective papers and is provided only for reference; it
is neither ranked against the lower block nor used to support the claims in
Table~\ref{tab:main_results}. Although some benchmark names overlap, the
published methods use author-specific training corpora, test constructions,
prompts, image preprocessing, coordinate conventions, and output parsers.
Consequently, matching benchmark names do not establish identical sample
identities, counts, or evaluation conditions.

We cannot report a verified unified RefGeo rerun for these five methods.
GeoViS depends on a VisRAG checkpoint that is not specified in its released
repository, while the available SkyMoE release does not provide a runnable
implementation and checkpoint. GeoChat, EarthGPT, and SkySenseGPT have partial
code or checkpoints available, but their released evaluation stacks do not
provide a validated end-to-end adapter for the exact six-source RefGeo JSONL,
prompt, image preprocessing, coordinate conversion, and strict parsing
contract used here. An ad hoc conversion would confound model quality with
integration choices, so we do not present such runs as comparable RefGeo
measurements. The missing cells and Overall values in the upper block are
therefore left as \na, and these values are excluded from the main table.

\begin{figure}[!t]
\centering
\includegraphics[width=\columnwidth]{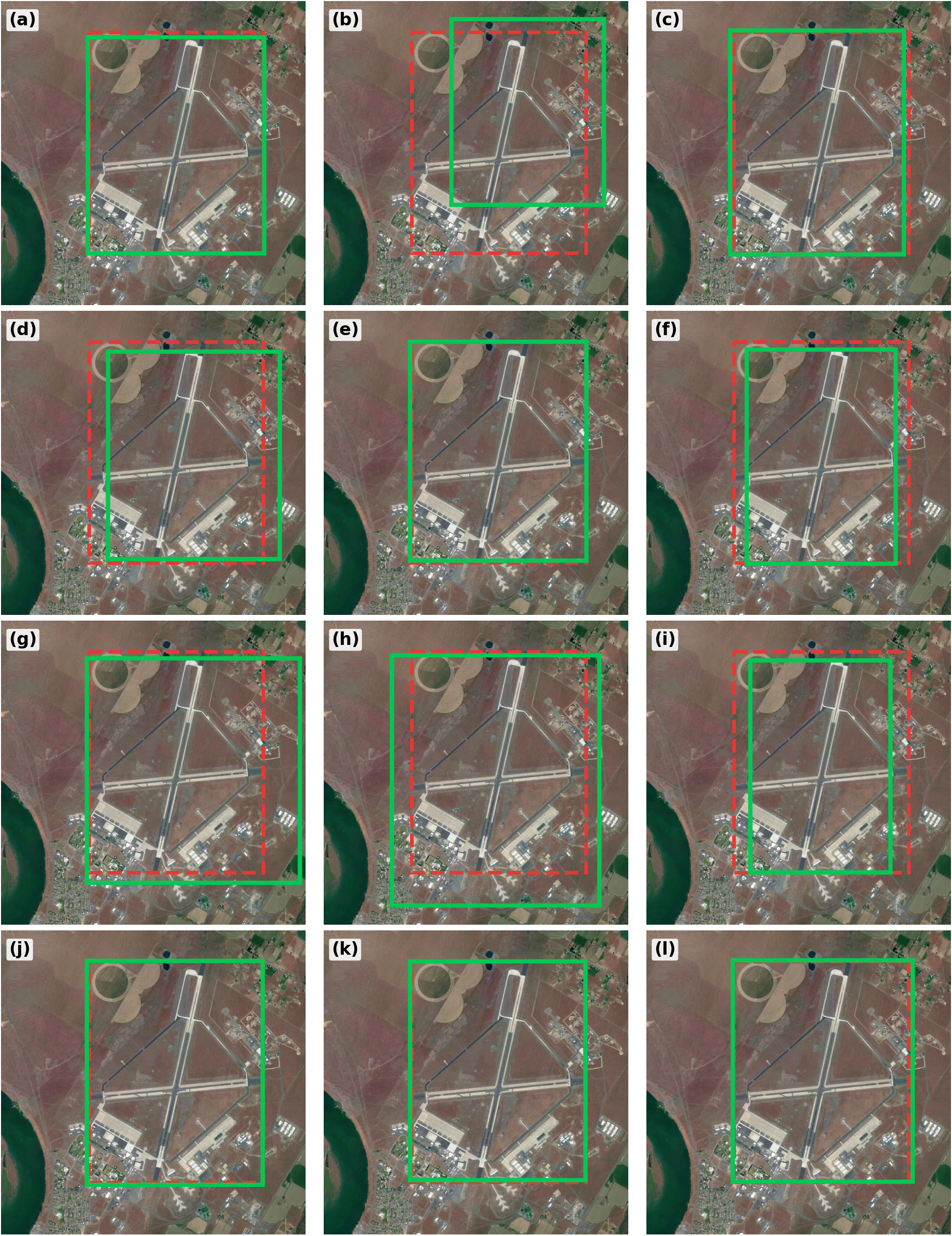}
\caption{Supported-query outputs: (a) Qwen3-VL-2B, (b) LLaVA-7B, (c,d) their fine-tuned variants, (e) GeoGround, (f) Geo-R1-3B, (g) RemoteReasoner, (h) GeoPixel, (i) Text4Seg, (j) Qwen3-VL-2B APD-T, (k) GeoGround APD-T, and (l) Qwen3-VL-2B APD-S. Dashed red is ground truth; solid green predictions have IoU$>0.5$.}
\label{fig:external_supported_outputs}
\end{figure}

\begin{figure}[!t]
\centering
\includegraphics[width=\columnwidth]{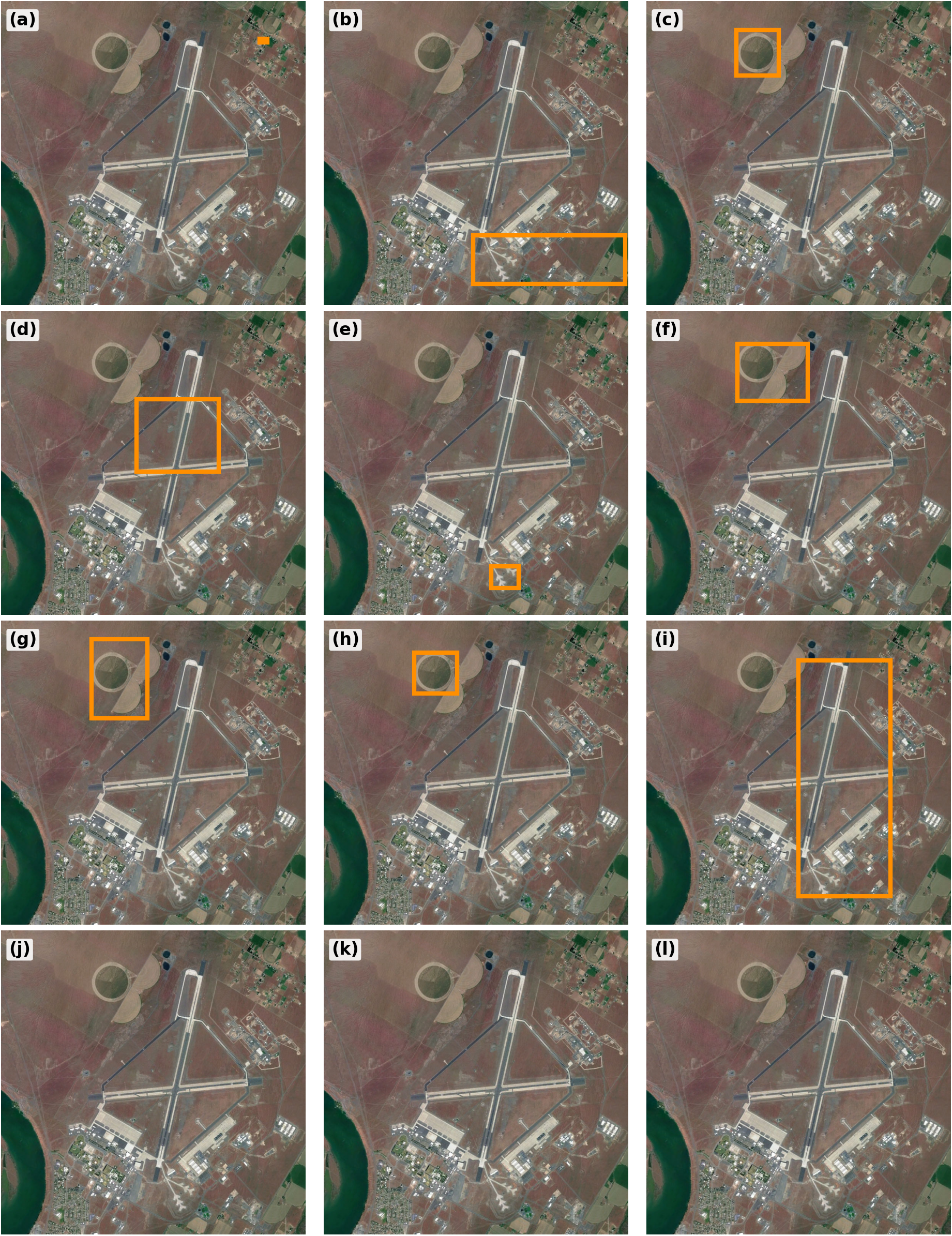}
\caption{Far-Category outputs; panels follow Figure~\ref{fig:external_supported_outputs}. Methods (a)--(i) return boxes, whereas APD variants (j)--(l) return \absentresp{}.}
\label{fig:external_absent_outputs}
\end{figure}

\subsection{Repeated-Prediction Examples}

\subsection{Representative Model Inputs and Outputs}

Figures~\ref{fig:external_supported_outputs} and
\ref{fig:external_absent_outputs} include every Table~\ref{tab:main_results}
method for which complete paired per-sample predictions are available. Both
figures use a DIOR-RSVG sample. The supported input is ``The orange and
gray airport in the middle,'' with ground truth
\bboxresp{[[230,82,689,662]]}\texttt{</bbox>}; the paired Far-Category input is ``Locate the
solar farm.'' For this input, no ground-truth box exists.

Thus, the common supported example confirms that every evaluated method can
perform localization, while the paired absent-target example separates this
ability from calibrated rejection. Cross-protocol methods without
verified paired per-sample prediction artifacts are not visualized. Exact normalized boxes, APD decisions, and source-file hashes are retained in
the qualitative manifest.

\subsection{APD-T Configuration and Metrics}

APD-T constructs the CIoU consistency graph and summarizes its dominant component by the coordinate-wise median. The remaining thresholds are $\tau_{dis}=0.05$ and $\tau_{ov}=0.5$. The VLM-specific $\eta$ values, calibrated from paired Original and Far-Category Counterfactual stability statistics and reused for Near-Category routing, are 1.02/1.67/1.40/1.06/1.04 for GeoGround, base/fine-tuned LLaVA-7B, and base/fine-tuned Qwen3-VL-2B.

Pooled scores are sample-count weighted; all comparisons use the same data order and parser.

\end{document}